%% file: iclr2021_conference.tex
\documentclass[11pt, other options]{article} 
\usepackage{iclr2021_conference,times}

\input{math_commands.tex}

\usepackage{hyperref}
\usepackage{url}
\input{latexdefs.tex}

\usepackage{wrapfig}
\usepackage{lipsum}

\makeatletter
\newcounter{phase}[algorithm]
\newlength{\phaserulewidth}
\newcommand{\setphaserulewidth}{\setlength{\phaserulewidth}}

\makeatother

\setphaserulewidth{.7pt}

\newlength{\whilewidth}
\algdef{SE}[parWHILE]{parWhile}{EndparWhile}[1]
  {\parbox[t]{\dimexpr\linewidth-\algmargin}{%
     \hangindent\whilewidth\strut\algorithmicwhile\ #1\ \algorithmicdo\strut}}{\algorithmicend\ \algorithmicwhile}%
\algnewcommand{\parState}[1]{\State%
  \parbox[t]{\dimexpr\linewidth-\algmargin}{\strut #1\strut}}

\usepackage[compact]{titlesec}
\titlespacing{\section}{0pt}{2ex}{1ex}
\titlespacing{\subsection}{0pt}{1ex}{0ex}
\titlespacing{\subsubsection}{0pt}{0.5ex}{0ex}
\title{One-Shot Neural Architecture Search \\ Via Compressive Sensing}

\author{Minsu Cho\textsuperscript{1}, Mohammadreza Soltani\textsuperscript{2}, and Chinmay Hegde\textsuperscript{1}\\
\textsuperscript{1} New York University, \textsuperscript{2} Duke University \\
\texttt{mc8065@nyu.edu, mohammadreza.soltani@duke.edu, chinmayh@nyu.edu} \\
}

\iclrfinalcopy 
\begin{document}

\maketitle

\begin{abstract}
Neural Architecture Search remains a very challenging meta-learning problem. Several recent techniques based on parameter-sharing idea have focused on reducing the NAS running time by leveraging proxy models, leading to architectures with competitive performance compared to those with hand-crafted designs. In this paper, we propose an iterative technique for NAS, inspired by algorithms for learning low-degree sparse Boolean functions. We validate our approach on the DARTs search space~\citep{liu2018DARTs} and NAS-Bench-201~\citep{Yang2020NAS}. In addition, we provide theoretical analysis via upper bounds on the number of validation error measurements needed for reliable learning, and include ablation studies to further in-depth understanding of our technique. 
\end{abstract}



\section{Introduction}
\label{sec:intro}

The choice of a suitable neural network architecture for complex prediction tasks such as image classification often requires substantial trial-and-error. Recently, there has been a growing interest to \emph{automatically} select the architecture of neural networks that can achieve competitive (or better) results over hand-designed architectures. Neural architecture search (NAS) tries to automate this hand-design process by constructing competitive architectures with as small computational budgets as possible ~\citep{zoph2018learning, liu2018progressive, liu2018DARTs, xie2018snas, bender2018understanding, li2019random}. 


In a departure from traditional methods, we approach the NAS problem via the lens of \emph{compressive sensing}. The field of compressive sensing (or sparse recovery), introduced by the seminal works of \citet{candes2004robust,donoho2006compressed}, has received significant attention in ML community over the last decade and has influenced advances in nonlinear and combinatorial optimization. Here, we develop a new NAS method called CoNAS (Compressive sensing-based Neural Architecture Search), which merges ideas from sparse recovery with the so-called ``one-shot'' architecture search methods~\citep{bender2018understanding, li2019random}. Our contribution is twofold. First, CoNAS uses a new \emph{search space} that permits exploration of a large(r) number of diverse candidate architectures. Second, it utilizes a new \emph{search strategy} that borrows ideas from the recovery of Boolean functions from their (sparse) Fourier expansions. We show how a combination of these two ideas leads to improved NAS performance.

\section{CoNAS: A new NAS approach}
\label{sec: alg}

\noindent\textbf{Overview.} Our proposed algorithm, Compressive sensing-based Neural Architecture Search (CoNAS), combines ideas from learning a sparse graph (Boolean Fourier analysis) and one-shot NAS. As we mentioned above, CoNAS consists of two components: a newly defined search space, and a more practical search strategy. Before discussing these two parts, we review some basics.

\paragraph{Fourier analysis of Boolean functions.}
We follow the treatment given in~\cite{o2014analysis}. A {real-valued Boolean function} is one that maps $n$-bit binary vectors (i.e., vertices of a hypercube) to a real number: $f : \{-1,1\}^n \rightarrow \mathbb{R}$. Such functions can be represented in a basis comprising real multilinear polynomials called the \emph{Fourier} basis, defined as follows. (We denote the vectors with bold letters. Also, $[n]$ denotes the set $\{1,2\dots,n\}$. Hence, the power set of $[n]$ is denoted by $2^{[n]}$.
\begin{definition}
\label{def:fourier basis}
For $S \subseteq [n]$, define the parity function $\chi_S: \{-1, 1\}^n \rightarrow \{-1, 1\}$ such that $\chi_S(\bm{\alpha}) = \prod_{i \in S} \alpha_i$. Then, the Fourier basis is the set of all $2^n$ parity functions $\{\chi_S\}$.
\end{definition}
The key fact is that the basis of parity functions forms an $K$-bounded orthonormal system (BOS) with $K=1$, therefore satisfying two properties: 
\begin{align}
\label{eq:BOS}
\langle \chi_S, \chi_T \rangle = 
\begin{cases}
    1, & \text{if } S = T\\
    0, & \text{if } S \neq T
\end{cases}
\:\:\:\:\:\:\:\:\:\:\: \textrm{and} && \sup_{\bm{\alpha} \in \{-1, 1\}^n} |\chi_S(\bm{\alpha})| \leq 1 \:\:\: \textrm{for all }S \subseteq [n],
\end{align}
Due to the orthonormality, any Boolean function $f$ has a unique Fourier representation, given by $f(\bm{\alpha}) = \sum_{S \subseteq [n]} \hat{f}(S) \chi_S(\bm{\alpha})$, with Fourier coefficients $\hat{f}(S) = \mathbb{E}_{\bm{\alpha}\in\{-1,1\}^n}[f(\bm{\alpha})\chi_S(\bm{\alpha})]$ where expectation is taken with respect to the uniform distribution over the vertices of a hypercube.

\begin{wrapfigure}{R}{0.6\textwidth}
\begin{minipage}{0.6\textwidth}
    \vspace{-2.3em}
    \begin{algorithm}[H]
    \caption{Pseudocode: CoNAS}\label{alg:conas}
    \begin{algorithmic}
        \State \textbf{Inputs:} Number of measurements $m$, number of coefficients $s$, lasso parameter $\lambda$
        \State Train one-shot model ($f$) with parameter sharing (RSPS)
        \While{not enough measurements ($k < m$)}
            \State Randomly sample a sub-architecture binary encoded vector $\alpha_k$
            \State Collect test loss, $\rvy = \{L(f(\alpha_1)), \ldots, L(f(\alpha_k))\}$
            \State $k \gets k+1$
        \EndWhile
        \State Construct graph sampling matrix $\rmA$ from $\{\alpha_1,\ldots,\alpha_m\}$
        \State Solve $\rvu^* = \argmin_{\rvu} \|\rvy - \rmA\rvu\|_2^2 + \lambda\|\rvu\|_1$. 
        \State Approximate $g\approx f$ with $s$ absolutely largest coefficients from $\rvu^*$
        \State Compute minimizer $\rvz = \argmin_{\alpha}g(\alpha)$ via brute force.
        \State \Return Constructed cell from $\rvz$
    \end{algorithmic}
    \end{algorithm}
    \vspace{-3em}
\end{minipage}
\end{wrapfigure}


A modeling assumption is that the Fourier spectrum of the function is concentrated on monomials of small degree ($\leq d$). This corresponds to the case where $f$ is a decision tree~\citep{hazan2017hyperparameter}, and allows us to simplify the Fourier expansion by limiting its support. Let $\mathcal{P}_d \subseteq 2^{[n]}$ be a fixed collection of Fourier basis 
such that 
$\mathcal{P}_d \coloneqq \{\chi_S : S \subseteq 2^{[n]}, |S| \leq d\}$
Then $\mathcal{P}_d$ induces a function space consisting of all functions of order $d$ or less, denoted by $\mathcal{H}_{\mathcal{P}_d} := \{f : Supp[\hat{f}] \subseteq \mathcal{P}_d\}$. For example, $\mathcal{P}_2$ allows us to express the function $f$ with at most $\sum_{l=0}^{d} \binom{n}{l} \equiv \mathcal{O}(n^2)$ Fourier coefficients.  

Lastly, if we have a prior knowledge of some set of bits ${J}$, we use an operation called \emph{restriction}.
\begin{definition}
\label{def:restriction}
Let $f: \{-1, 1\}^n \rightarrow \mathbb{R}$, $(J, \overline{J})$ be a partition of $[n]$, and $z \in \{-1, 1\}^{\overline{J}}$. The restriction of $f$ to $J$ using $z$ denoted by $f_{J|z}: \{-1, 1\}^J \rightarrow \mathbb{R}$ is the subfunction of $f$ given by fixing the coordinates in $\overline{J}$ to the bit values $z$.
\end{definition}

We now discuss two main components of CoNAS.

\noindent\textbf{1-Search Space.} Following the approach of DARTs~\citep{liu2018DARTs}, we define a directed acyclic graph (DAG) with all predecessor nodes are connected to every intermediate node with all possible operations. We represent any sub-graph of the DAG using a binary string $\bm{\alpha}$ called the \emph{architecture encoder}. Its length is the total number of edges in the DAG, excluding the edges connect to the output node. A $1$ (resp. $-1$) in $\bm{\alpha}$ indicates an active (resp. inactive) edge.

\begin{figure}[!t]
\centering
    \resizebox{0.92\linewidth}{!}{
    \input{diagram.tex}}
    \vspace{1em}
    \caption{\sl Diagram inspired by~\citet{bender2018understanding}. The example architecture encoder $\bm{\alpha}$ samples a sub-architecture for $N=5$ nodes (two intermediate nodes) with five different operations. Each component in $\bm{\alpha}$ maps to the edges one-to-one in all \emph{Choice} blocks in a cell. 
    Since the CNN search space finds both \emph{normal cell} and \emph{reduce cell}, the length of $\bm{\alpha}$ is equivalent to $(2+3) \cdot 5 \cdot 2 = 50$.} 
    \label{fig:diagram}
    \vspace{.3cm}
\end{figure}

Figure~\ref{fig:diagram} gives an example of how the architecture encoder $\bm{\alpha}$ samples the sub-architecture of the fully-connected model in the case of a convolutional neural network. The goal of CoNAS is to find the ``best'' encoder $\bm{\alpha^*}$, which is "close enough" to the best achievable validation accuracy by constructing the final model with $\bm{\alpha^*}$ encoded sub-graph.


\paragraph{2-Search Strategy.}
We propose a compressive measuring strategy to approximate the one-shot model with a Fourier-sparse Boolean function. Let $f: \{-1, 1\}^n \rightarrow \R$ map the sub-graph of the one-shot pre-trained model encoded by $\bm{\alpha}$ to its validation performance. We collect a small number of function evaluations of $f$ and reconstruct the Fourier-sparse function $g \approx f$ via a sparse recovery algorithm with randomly sampled measurements (test loss).  Then, we solve $\argmin_{\bm{\alpha}}g(\bm{\alpha})$ by exhausting enumeration over all coordinates, supports of the Boolean function $g$. We obtain a cell to construct the final architecture by activating the edges such that $\alpha_i^* = 1$. If the solution does not return enough edges, restricting the hypercube allows the iterative process to find more edges. From Definition~\ref{def:restriction}, we restrict the approximate function $g$ by fixing bit values found in the previous solution and repeat the sparse recovery by randomly sampling sub-graph in the remaining edges. 


\paragraph{Full Algorithm.}
We now describe CoNAS in detail, with pseudocode provided in Algorithm~\ref{alg:conas}.

We first train a parameter-shared one-shot model~\citep{li2019random} with standard backpropagation; however, we only update the weights corresponding to the randomly sampled sub-graph edges for each minibatch. 
Then, we randomly sample sub-graphs by generating architecture encoder strings $\bm{\alpha} \in \{-1, 1\}^n$ from some distribution. (e.g., a $Bernoulli(p)$ distribution for each bit of $\bm{\alpha}$ independently).  

In the second stage, we collect $m$ measurements of randomly sampled sub-architecture performance denoted by $\mathbf{y} = (L(f(\bm{\alpha}_1)), L(f(\bm{\alpha}_2)), \ldots, L(f(\bm{\alpha}_m)))^T$.  Next, we construct the \emph{graph-sampling matrix} $\mathbf{A} \in \{-1, 1\}^{m \times |\mathcal{P}_d|}$ with entries 
\begin{align}
\label{eq:sampling matrix}
\mathbf{A}_{l, k} = \chi_{S_k}(\bm{\alpha}_l), \:\:\:\:\:\:\: l \in [m], k \in [|\mathcal{P}_d|], S \subseteq [n], |S| \leq d,
\end{align}
where $d$ is the maximum degree of monomials in the Fourier expansion, and $S_k$ is the index set corresponding to the k\textsuperscript{th} Fourier basis. We solve the familiar Lasso problem~\citep{tibshirani1996regression}:
\begin{align}
\label{eq:Lasso}
    \mathbf{u}^* = \argmin_{\mathbf{u} \in \mathbb{R}^{|\mathcal{P}_d|}} \|\mathbf{y} - \mathbf{A} \mathbf{u}\|_2^2 + \lambda\|\mathbf{u}\|_1, 
\end{align}
to (approximately) recover the global optimizer $\rvu^*$, the vector contains the Fourier coefficients corresponding to $\mathcal{P}_d$. We define an approximate function $g \approx f$ with Fourier coefficients with the top-$s$ (absolutely) largest coefficients from $\rvu^*$, and solve $\bm{\alpha}^* = \argmin_{\bm{\alpha}} g(\bm{\alpha})$ by computing all the possible points (brute force) in the subcube defined by the support of $g$ (this computation is feasible if $s$ is small). 
Finally, we obtain a cell to construct the final architecture by activating the edges corresponding to all ${i \in [n]}$ such that $\alpha_i^* = 1$. 

\paragraph{Theoretical support for CoNAS.} We first note that a system of linear equations given by $\mathbf{y} = \mathbf{Au}$ with the graph-sampling matrix $\mathbf{A} \in \{-1, 1\}^{m \times O(n^d)}$, measurements $\mathbf{y} \in \mathbb{R}^m$, and Fourier coefficient vector $\mathbf{u} \in \mathbb{R}^{O(n^d)}$ is an ill-posed problem when $m \ll O(n^d)$ for large $n$. However, if the graph-sampling matrix satisfies \emph{Restricted Isometry Property (RIP)}, the sparse coefficients, $\mathbf{u}$ can be uniquely recovered.
\begin{definition}
\label{def: rip}
A matrix $\mathbf{A}\in\mathbb{R}^{m\times \mathcal{O}(n^d)}$ satisfies the restricted isometry property of order $s$ with some constant $\delta$ if for every $s$-sparse vector $\mathbf{u}\in\mathbb{R}^{\mathcal{O}(n^d)}$ (i.e., only $s$ entries are non-zero) the following holds:
$$(1-\delta) \|\mathbf{u}\|_2^2\leq\|\mathbf{Au}\|_2^2\leq(1+\delta) \|\mathbf{u}\|_2^2.$$
\end{definition}

To the best of our knowledge, the best known result with mild dependency on $\delta$ (i.e., $\delta^{-2}$) is due to~\cite{haviv2017restricted}, which we can apply for our setup. It is easy to check that the graph-sampling matrix $\mathbf{A}$ in our proposed CoNAS algorithm satisfies BOS for $K=1$ (\eqref{eq:sampling matrix}). We defer the proof in Appendix~\ref{appendix: cs}.

\begin{theorem}
\label{thm:recovery on main}
Let the graph-sampling matrix $A\in\{-1,1\}^{m\times \mathcal{O}(n^d)}$ be constructed by taking $m$ rows (random sampling points) uniformly and independently from the rows of a square matrix $\mathbf{M}\in\{-1,1\}^{\mathcal{O}(n^d)\times \mathcal{O}(n^d)}$. Then the normalized matrix $\mathbf{A}$ with $m = \mathcal{O}(\log^2(\frac{1}{\delta})\delta^{-2}s\log^2(\frac{s}{\delta})d \log(n))$ with probability at least $1-2^{-\Omega(d\log n\log(\frac{s}{\delta}))}$ satisfies the restricted isometry property of order $s$ with constant $\delta$; as a result, every $s$-sparse vector $\mathbf{u}\in \mathbb{R}^{\mathcal{O}(n^d)}$ can be recovered from the sample $y_i$'s $\mathbf{y} = \mathbf{Au} = \big(\sum_{j=1}^{|\mathcal{O}(n^d)|} u_j\mathbf{A}_{i,j}\big)_{i=1}^{m}$ by solving the LASSO problem in (equation~\ref{eq:Lasso}).
\end{theorem}



In essence, Theorem~\ref{thm:recovery on main} provides a successful guarantee for recovering the optimal sub-network of a given size given a sufficient number of performance measurements. 

\section{Experiments and Results}
\label{sec:result}
In this section, we illustrate the efficacy of our proposed approach through some experimental results. We start by experimenting on two popular NAS benchmarks: (i) a CNN search on CIFAR-10 on DARTs~\citep{liu2018DARTs} search space, (ii) a CNN search on NAS-Bench-201~\citep{Yang2020NAS} search space. We compare CoNAS with the following NAS algorithms in DARTs search space: NASNet~\citep{zoph2018learning}, 
SNAS~\citep{xie2018snas}, ENAS~\citep{pham2018efficient}, DARTs~\citep{liu2018DARTs}, random search from \cite{liu2018DARTs}, and RSPS~\citep{li2019random}. Our evaluation setup for training the final architecture is the same as one reported by DARTs and NAS-Bench-201. We defer the search/evaluation setup details and ablation studies to the section~\ref{appendix: Hyperparameter Setup, } in the Appendix. 
The implementation of CoNAS is available from \href{https://github.com/chomd90/CoNAS\_release}{\textcolor{blue}{this link\footnote{https://github.com/chomd90/CoNAS\_release}}}.



\begin{table}[t]
    \centering
    \caption{\sl \textbf{Comparison with existing NAS methods on CIFAR-10 DARTs search space}} \vspace{1em}
    \label{table: CIFAR10 Comparison Table}
    \begin{tabular}{c | c | c | c | c }
        \hline
        \multicolumn{1}{l}{\textbf{Method}} & Test Error (\%) & Params (M) & Multi-Add (M) & Search (GPU day)\\
        \hline
        \multicolumn{1}{l}{NASNet-A} & $2.65$ & 3.3 & - & 2000 \\
        \multicolumn{1}{l}{GDAS} & $2.82$ & 2.5 & - & 0.17 \\ 
        \multicolumn{1}{l}{SNAS} & $2.85 \pm 0.02$ & 2.8 & - & 1.5  \\
        \multicolumn{1}{l}{ENAS}& 2.89 & 4.6 & - & 0.45 \\
        \multicolumn{1}{l}{DARTs} & $2.76\pm0.09$ & 3.3 & 548 & 4 \\
        \multicolumn{1}{l}{Random Search} & $3.29 \pm 0.15$ & 3.1 & - & 4 \\
        \multicolumn{1}{l}{RSPS} & $2.71|2.85\pm0.08$ & 3.7 & 634 & 2.7 \\
        \hline
        \multicolumn{1}{l}{CoNAS (1)} & $2.59|2.67\pm0.06$ & 3.2 & 450 & 0.35 \\
        \multicolumn{1}{l}{CoNAS (2)} & $2.73|2.85\pm0.08$ & 2.7 & 430 & 0.35 \\
        \hline
    \end{tabular}
\end{table}

\noindent \textbf{DARTs Search Space} We run CoNAS in two different settings: (1) cells from the optimization problem~(\eqref{eq:Lasso}) and (2) the sub-graph from (1) to strictly match DARTs cell by manually dropping some operations (up to 2 edges for each intermediate nodes). We conduct the final training with exact hyperparameter setups used in DARTs. The average test error of our experiment uses five random seeds. 
Figure~\ref{appendix: cells} in the Appendix includes the cells found from CoNAS. Our approach achieves the test error of $2.59\%$ surpassing the RSPS, DARTs, and random search.


\noindent \textbf{NAS-Bench-201 Search Space} We evaluate our algorithm on NAS-Bench-201, a NAS benchmark designed for all cell-based NAS methods. We implement our algorithm on top of the existing Random Search with Parameter Sharing (RSPS) implementation in the NAS-Bench-201 library. 
We randomly sample sub-graphs from a trained one-shot model and solve the optimization problem~(\eqref{eq:Lasso}).
If the solution picks more than one operation (an edge) between the nodes, we randomly choose an operation uniformly from the given solution. The average test error on NAS-Bench-201 uses three different random seeds.
Table~\ref{table: NAS-Bench-201 Table} compares cell-based NAS methods on NAS-Bench-201. We observe that both approaches consistently find a better cell structure than RSPS. 

\begin{table}[t]
    \centering
    \setlength{\tabcolsep}{2.0pt}
    \vspace{1.5em}
    \caption{\sl \textbf{Comparison with existing NAS methods on NAS-Bench-201}} \vspace{1em}
    \label{table: NAS-Bench-201 Table}
    \begin{tabular}{c | c | c | c | c | c | c | c}
        \hline
        \multirow{2}{*}{\textbf{Method}} & \multicolumn{1}{c|}{Search} & \multicolumn{2}{c|}{CIFAR-10} & \multicolumn{2}{c|}{CIFAR-100} & \multicolumn{2}{c}{ImageNet-16-120} \\
        \cline{2-8}
        & (seconds) & validation & test & validation & test & validation & test \\ 
        \hline
        DARTs & 35781.80 & 39.77±0.00 & 54.30±0.00 & 15.03±0.00 & 15.61±0.00 & 16.43±0.00 & 16.32±0.00 \\
        GDAS &  31609.80 & 89.89±0.08 & 93.61±0.09 & 71.34±0.04 & 70.70±0.30 & 41.59±1.33 & 41.71±0.98 \\
        ENAS & 14058.80 & 37.51±3.19 & 53.89±0.58 & 13.37±2.35 & 13.96±2.33 & 15.06±1.95 & 14.84±2.10 \\
        RSPS & 8007.13 & 80.42±3.58 & 84.07±3.61 & 52.12±5.55 & 52.31±5.77 & 27.22±3.24 & 26.28±3.09 \\ 
        \hline
        CoNAS & 8222.80 & 88.40±2.79 & 91.22±3.08 & 65.82±5.72 & 66.39±5.51 & 39.51±6.95 & 38.82±7.01 \\
        \hline
    \end{tabular}
\end{table}


\newpage

\section*{Acknowledgements}

MC and CH are supported in part by NSF grants CCF-2005804 and CCF-1815101.

\bibliography{iclr2021_conference}
\bibliographystyle{iclr2021_conference}

\appendix
\section{Appendix}

\subsection{Search/Evaluation Protocols and Hyperparameter Setup}
\label{appendix: Hyperparameter Setup}

We conduct the exact equivalent evaluation training protocol of DARTs for the fair comparison. We adopt a NAS-Bench-201 API for the NAS-Bench-201 search space, which provides all available cells' precomputed performance, including test loss and accuracy. 

\noindent\textbf{Search Protocol in DARTs Search Space} We train a one-shot architecture equivalent to RSPS with a cell containing $N=7$ nodes with two nodes as inputs and one node as output. We randomly sample sub-graphs by generating architecture encoder strings $\alpha \in \{-1, 1\}^n$ using a $Bernoulli(p)$ distribution for each bit of $\alpha$ independently (we set $p=0.5$). We set the threshold for gathering measurements (loss) to remove abnormally high loss values. We use seven operations: $3\times3$ and $5\times5$ separable convolutions, $3\times3$ and $5\times5$ dilated convolutions, $3\times3$ average pooling, $3\times3$ max pooling, and Identity. We equally divide the 50,000-sample training set into training and validation sets, following \citet{li2019random} and \citet{liu2018DARTs}. 

\noindent\textbf{Search Protocol in NAS-Bench-201 Search Space} NAS-Bench-201 provides the predefined search space with four operations (five operations including a zero operation) and four nodes. We uniformly random sample the sub-graph from all possible cells in the search space. 

Table~\ref{appendix: search DARTs hp} and Table~\ref{appendix: search bench hp} include the training protocol and hyperparameters used in CoNAS on DARTs search space and NAS-Bench-201, respectively. 

\begin{table}[H]
\vspace{0.5em}
\setlength{\tabcolsep}{2.0pt}
    \begin{minipage}[b]{0.45\linewidth}\centering
    \caption{DARTs search space searching hyperparameter set.}
    \vspace{1.0em}
    \label{appendix: search DARTs hp}
    \begin{tabular}{c | c || c | c}
        \hline
        optimizer & SGD & initial LR & 0.025 \\
        Nesterov & Yes & ending LR & 0.001 \\
        momentum & 0.9 & LR schedule & cosine \\
        weight decay & 0.0003 & epoch & 100 \\
        batch size & 64 & initial channel & 16 \\
        cells \# & 8 & cutout & No \\
        ops \# & 7 & nodes \# & 7 \\
        random flip & p=0.5 & random crop & Yes \\
        normalization & Yes & threshold & 1.5 \\
        measurements & 3000 & lasso ($\lambda$) & 10.0 \\
        coefficient \# ($s$) & 12 & degree ($d$) & 2 \\
        \hline
    \end{tabular}
    \end{minipage}
    \hspace{1.5em}
    \begin{minipage}[b]{0.45\linewidth}\centering
    \caption{NAS-Bench-201 search space searching hyperparameter set.}
    \vspace{1.0em}
    \label{appendix: search bench hp}
    \begin{tabular}{c | c || c | c}
        \hline
        optimizer & SGD & initial LR & 0.025 \\
        Nesterov & Yes & ending LR & 0.001 \\
        momentum & 0.9 & LR schedule & cosine \\
        weight decay & 0.0005 & epoch & 100 \\
        batch size & 64 & initial channel & 16 \\
        cells \# & 5 & cutout & No \\ 
        ops \# & 4 & nodes \# & 4 \\
        random flip & p=0.5 & random crop & Yes \\
        normalization & Yes & threshold & 1.8 \\
        measurements & 300 & lasso ($\lambda$) & 5.0 \\
        coefficient \# ($s$) & 14 & degree ($d$) & 2 \\
        \hline
    \end{tabular}
    \end{minipage}
\end{table}

\noindent \textbf{Evaluation Protocol in DARTs Search Space} We evaluate the cell found from CoNAS with exactly equivalent training protocol to DARTs. We use either NVIDIA Quadro RTX 8000 or NVIDIA TITAN RTX for the final architecture training process (1 GPU training). Table~\ref{appendix: final protocol1} shows the final evaluation protocol details. We set random seeds from 0 to 4 for the average test error on Table~\ref{table: CIFAR10 Comparison Table}. 

\begin{table}[H]
    \centering
    \setlength{\tabcolsep}{2.0pt}
    \vspace{1.5em}
    \caption{DARTs search space final evaluation hyperparameter set.} \vspace{1em}
    \label{appendix: final protocol1}
    \begin{tabular}{c | c || c | c}
        \hline
        optimizer & SGD & initial LR & 0.025 \\
        Nesterov & Yes & ending LR & 0 \\
        momentum & 0.9 & LR schedule & cosine \\
        weight decay & 0.0003 & epoch & 600 \\
        batch size & 96 & initial channel & 36 \\
        random flip & p=0.5 & random crop & Yes \\
        normalization & Yes & cutout & Yes \\
        drop-path & 0.2 & cutout length & 16 \\
        cells \# & 20 & aux weight & 0.4 \\
        grad clip & 5.0 & parallel training & No \\
        \hline
    \end{tabular}
\end{table}

\begin{figure}[t]
    \centering
    \renewcommand{\arraystretch}{0.2}
    \def\sw{0.40\linewidth}
    \begin{tabular}{c c}
        CoNAS-Normal (1) & CoNAS-Reduce (1) \\
        \includegraphics[width=\sw]{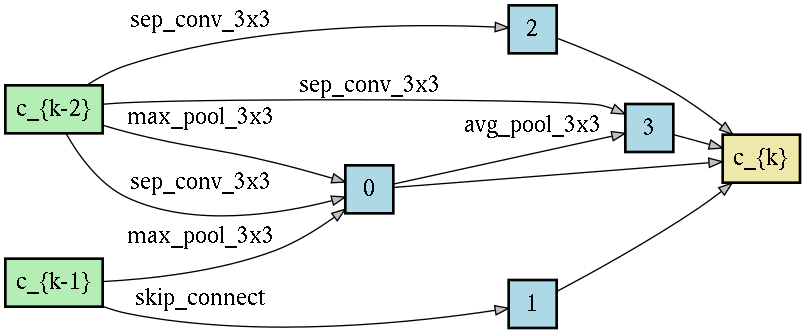} &
        \includegraphics[width=\sw]{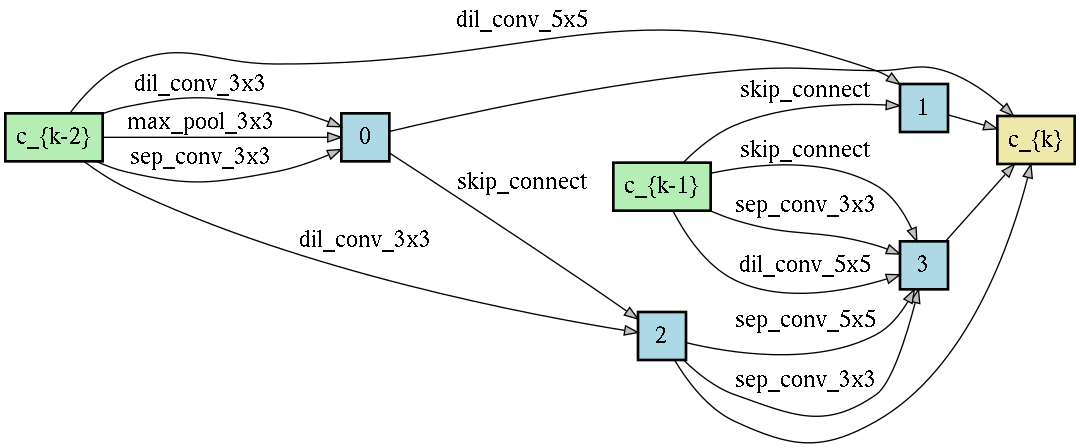} \\ \vspace{1.5em}
        \includegraphics[width=\sw]{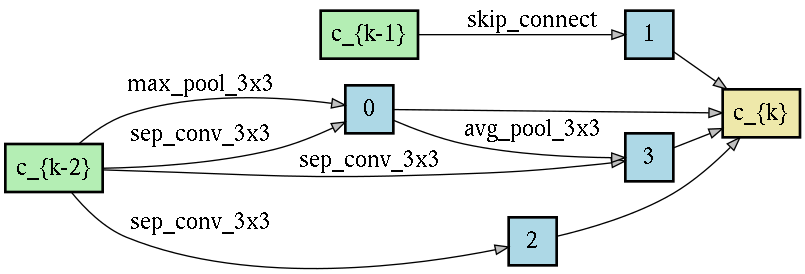} &
        \includegraphics[width=\sw]{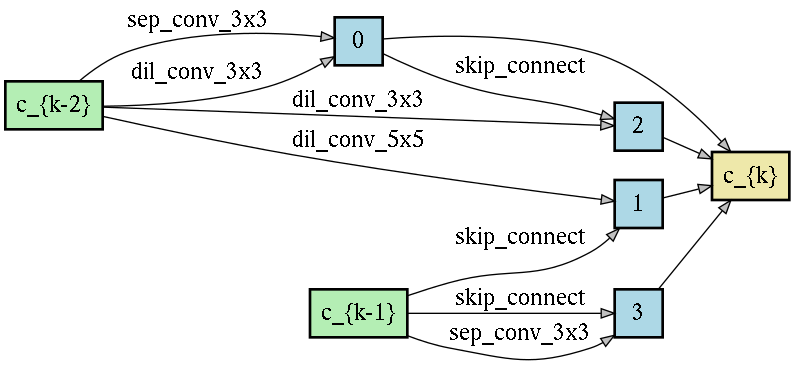} \\
        CoNAS-Normal (2) & CoNAS-Reduce (2) \\
    \end{tabular}
    \vspace{1.5em}
    \caption{The cell in the first (resp. second) row corresponds to the CoNAS(1) (resp. CoNAS(2)) results in Table~\ref{table: CIFAR10 Comparison Table}. CoNAS(2) is the sub-graph of CoNAS(1) which strictly meets the requirement for DARTs search space (only up to two edges allowed for intermediate nodes).}
    \vspace{1.5em}
    \label{appendix: cells}
\end{figure}

\subsection{NAS Literature}
\label{appendix: NAS literature}

\textbf{Neural Architecture Search.} Early NAS approaches used RL-based controllers~\citep{zoph2018learning, pham2018efficient}, evolutionary algorithms \citep{real2018regularized}, or sequential model-based optimization (SMBO)~\citep{liu2018progressive}, and showed competitive performance with manually-designed architectures such as deep ResNets~\citep{he2016deep} and DenseNets~\citep{huang2017densely}. However, these approaches required substantial computational resources, running into thousands of GPU-days. Subsequent NAS works have focused on boosting search speeds by proposing novel search strategies, such as differentiable search technique via gradient-based  optimization~\citep{cai2018proxylessnas,liu2018DARTs,xie2018snas} and random search via sampling sub-networks from a one-shot super-network~\citep{bender2018understanding, li2019random}. To the best of our knowledge, no NAS method yet reported has explored compressive sensing techniques. 

\textbf{Differentiable Neural Architecture Search (DARTs).} Our CoNAS approach can be viewed as a refinement to DARTs (\citep{liu2018DARTs}) which performs bilevel optimization by relaxing the (discrete) architecture search space to a differentiable search space via softmax operations. The choice of alternative optimization on differentiable multi-objective formulation substantially speeds up the search by orders of magnitude while achieving competitive performance compared to previous works~\citep{zoph2016neural, zoph2018learning, real2018regularized, liu2018progressive}. 

\textbf{One-Shot Neural Architecture Search.} \citet{bender2018understanding} provide an extensive experimental analysis on one-shot architecture search based on weight-sharing. \citet{bender2018understanding} statistically showed the correlation between the one-shot model (super-graph) and stand-alone model (sub-graph) through the experiments. \citet{li2019random} proposes simplified training procedures without stabilizing techniques (e.g., path dropout schedule on a direct acyclic graph (DAG) and ghost batch normalization) from \citet{bender2018understanding}. As the final performance of the discovered architecture heavily relies on hyperparameter settings, \citet{li2019random} exactly accords hyperparameters and data augmentation techniques to DARTs for their experiments. This combination of random search via one-shot models with weight-sharing provides the best competitive baseline results reported in the NAS literature. Our CoNAS approach improves upon these reported results.

\textbf{Learning Sub-Networks.} \citet{stobbe2012learning} propose learning sparse sub-networks from a small number of random cuts; they also leverage ideas from compressive sensing and provide theoretical upper bounds for successful recover. Our CoNAS approach is directly inspired from their seminal work. \textcolor{black}{However, we emphasize essential differences: while \citet{stobbe2012learning} emphasize \emph{linear} measurements, CoNAS takes a different perspective by focusing on measurements that map sub-networks to performance, which are fundamentally \emph{nonlinear}. Moreover, our theoretical bounds use better Fourier-RIP bounds, and lead to improved results in terms of measurement complexity.} 

\textbf{Hyperparameter optimization.}
Building upon the approach of \citet{stobbe2012learning}, \citet{hazan2017hyperparameter} develop a spectral approach called \emph{Harmonica} for hyperparameter optimization (HPO) by encoding hyperparameters as binary strings. CoNAS also follows the same path, albeit for NAS. While NAS and HPO are sister meta-learning problems, we emphasize that our focus is exclusively on NAS, while \citet{hazan2017hyperparameter} exclusively focus on HPO. 

Moreover, the techniques of \citet{hazan2017hyperparameter} cannot be directly applied to the NAS problem. We need to define our search space, encode our search problem in terms of Boolean variables, and propose how to gather measurements. All these are new to our paper: in particular, CoNAS proposes gathering measurements within tractable sampling time via top of RSPS, while Harmonica naively gathers the approximated measurements by training the model for each randomly sampled hyperparameter choice. Finally, Harmonica requires invocation of a baseline hyperparameter optimization method (such as random search, successive halving~\citep{jamieson2016non}, or Hyperband~\citep{li2017hyperband}), which CoNAS does not require.

\subsection{Theoretical support from compressive sensing}
\label{appendix: cs}

The system of linear equations $\rvy = \rmA\rvu$ with the graph-sampling matrix $\rmA \in \{-1, 1\}^{m \times O(n^d)}$, measurements $\rvy \in \R^m$, and Fourier coefficient vector $\rvu \in \R^{O(n^d)}$ is an ill-posed problem when $m << O(n^d)$ for large $n$. However the sparse coefficients $\rvu$ can be recovered if the graph-sampling matrix satisfies \emph{Restricted Isometry Property} from Definition~\ref{def: rip}. 

There has been significant research during the last decade in proving upper bounds on the number of rows of bounded orthonormal dictionaries (matrix $\mathbf{A}$) for which $\mathbf{A}$ is guaranteed to satisfy the restricted isometry property with high probability. One of the first BOS results was established by~\citet{candes2006near}, where the authors proved an upper bound scales as $\mathcal{O}(sd^6\log^6 n)$ for a subsampled Fourier matrix. While this result is seminal, it is only optimal up to some \textit{polylog} factors. In fact, the authors in chapter $12$ of \cite{foucart2017mathematical} have shown a necessary condition (lower bound) on the number of rows of BOS which scales as $\mathcal{O}(sd\log n)$. In an attempt to achieve to this lower bound, the result in~\cite{candes2006near} was further improved by~\cite{rudelson2008sparse} to $\mathcal{O}(sd\log^2s\log(sd\log n)\log n)$. Motivated by this result, \cite{cheraghchi2013restricted} has even reduced the gap further by proving an upper bound on the number of rows as $\mathcal{O}(sd\log^3 s\log n)$.  The best known available upper bound on the number of rows appears to be $\mathcal{O}(sd^2\log s\log^2n)$; however with worse dependency on the constant $\delta$, i.e., $\delta^{-4}$ (please see~\cite{bourgain2014improved}). To the best of our knowledge, the best known result with mild dependency on $\delta$ (i.e., $\delta^{-2}$) is due to~\cite{haviv2017restricted}, and is given by $\mathcal{O}(sd\log^2s\log n)$. We have used this result for proving Theorem~\ref{thm:recovery on main}. 

\begin{theorem}
Let the graph-sampling matrix $A\in\{-1,1\}^{m\times \mathcal{O}(n^d)}$ be constructed by taking $m$ rows (random sampling points) uniformly and independently from the rows of a square matrix $\mathbf{M}\in\{-1,1\}^{\mathcal{O}(n^d)\times \mathcal{O}(n^d)}$. Then the normalized matrix $\mathbf{A}$ with $m = \mathcal{O}(\log^2(\frac{1}{\delta})\delta^{-2}s\log^2(\frac{s}{\delta})d \log(n))$ with probability at least $1-2^{-\Omega(d\log n\log(\frac{s}{\delta}))}$ satisfies the restricted isometry property of order $s$ with constant $\delta$; as a result, every $s$-sparse vector $\mathbf{u}\in \mathbb{R}^{\mathcal{O}(n^d)}$ can be recovered from the sample $y_i$'s $\mathbf{y} = \mathbf{Au} = \big(\sum_{j=1}^{|\mathcal{O}(n^d)|} u_j\mathbf{A}_{i,j}\big)_{i=1}^{m}$ by LASSO (equation~\ref{eq:Lasso}).
\end{theorem}

\begin{proof}
First, we note that the graph-sampling matrix $A$ is a BOS matrix with $K=1$; hence, directly invoking Theorem 4.5 of \cite{haviv2016list} to our setting, we can see that matrix $A$ satisfies RIP. Now according to Theorem 1.1 of \cite{candes2008restricted}, letting $\delta<\sqrt{2} -1$, the $\ell_1$ minimization or LASSO will recover exactly the $s$ sparse vector $u$. For instance, in our experiments, we have selected $m=3000$ which is consistent with our parameters, $d=2, s=14, n=196$. 
\end{proof}

\subsection{Ranking Correlation of RSPS.}
\label{appendix: RSPS rank correlation}

\begin{wrapfigure}{R}{0.5\textwidth}
\begin{minipage}{0.5\textwidth}
\begin{figure}[H]
    \vspace{-3.2em}
    \centering
    \captionsetup{width=\textwidth}
    \includegraphics[width=\textwidth]{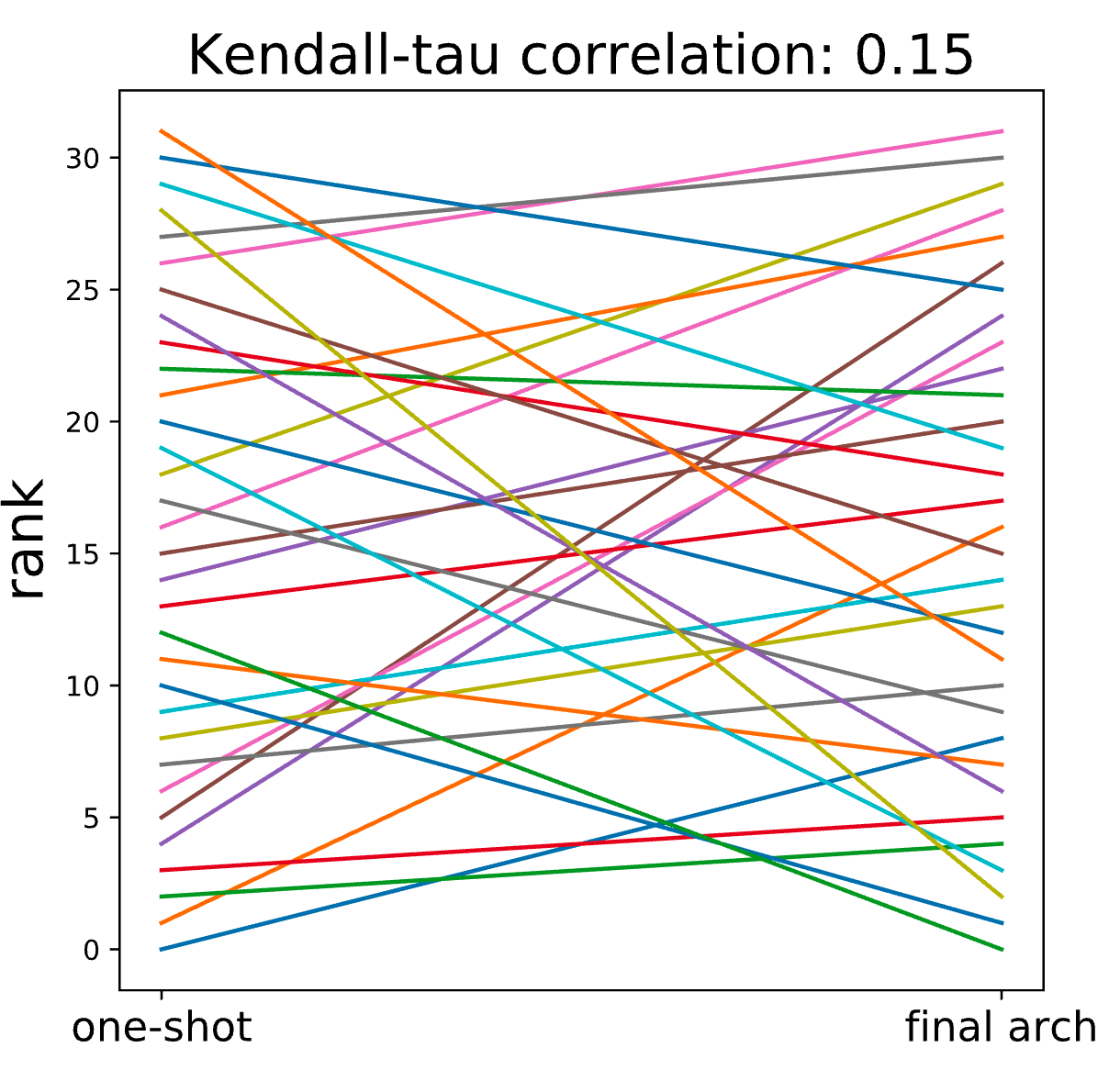}
    \vspace{0.5em}
    \caption{Ranking difference between one-shot model and actual architecture. The higher the Kendall-tau correlation, the more each line parallel to the x-axis.}
    \label{appendix: RSPS corr}
\end{figure}
\end{minipage}
\end{wrapfigure}

We leverage Random Search with Parameter Sharing (RSPS) to collect measurements quickly for sparse recovery with the Fourier basis of Boolean functions. 
We examine the correlation between performance estimations from parameter-shared models and performance from isolated training from scratch with concatenated cells
We first randomly sample 32 cells from a 100-epoch trained one-shot model (same search setup described in Appendix~\ref{appendix: Hyperparameter Setup}) with RSPS method on the CIFAR-10 dataset. Then we collect two sets of test losses to observe ranking correlation via Kendall tau correlation between one-shot model and macro-skeleton trained with 160 epochs for each randomly sampled cell. 
We matched the initial number of channels (16) and depth (8) of micro skeleton (equivalent to number of cells) to the one-shot model. We observed the correlation between one-shot model and macro-skeleton is weak with Kendall tau $\tau = 0.15$ with the two-sided p-value 0.23. 
Figure~\ref{appendix: RSPS corr} visually represents the rank correlation between one-shot model prediction and actual trained model in test loss. 
We point out that other factors that exist weakening the correlation, such as the same architecture training with different seeds, affect the ranking correlation~\citep{Yang2020NAS} and high variance of CIFAR-10 results even with exact training protocol~\citep{liu2018progressive}. 

\subsection{Stability on Lasso Parameters}
\label{appendix: lasso parameter}
We check our algorithm's stability on lasso parameter by observing the solution given exact same measurements. Denote $\alpha^*_{\lambda=l}$ as the architecture encoded output from CoNAS given $\lambda=l$. We compare the hamming distance and the test error between $\alpha^*_{\lambda=1}$ and other $\lambda$ values ($\lambda = 0.5, 2, 5, 10$). We train a parameter-shared one-shot network with five operations resulting the encoder length $(2+3+4+5)*5*2=140$. The average support of the solution from one sparse recovery is 15 out of the 140 length. The average hamming distance between two randomly generated binary strings with $\text{supp}(\alpha^*)=15$ from $100,000$ samples was $27.58\pm1.82$. Our experiment shows a stable performance under various lasso parameters with small hamming distances regards to various $\lambda$. Also we measure the average test error with 150 training epochs on different $\lambda$ values as shown in Table~\ref{table:lasso parameter testing}. For the baseline comparison, we compare CoNAS solutions with the randomly chosen architecture with 15 operations.

\begin{table}[h]
    \caption{Lasso Parameter Stability Experiment.}
    \label{table:lasso parameter testing}
    \vspace{1.0em}
    \centering
    \begin{tabular}{c c c c c c}
    \toprule
        Criteria & $\lambda = 0.5$ & $\lambda = 2.0$ & $\lambda = 5.0$ & $\lambda = 10.0$ & Random\\
    \midrule
        \multicolumn{1}{l}{Hamming Distance} & $0$ & $0$ & $8$ & $12$ & 29\\
        \multicolumn{1}{l}{Test Error (\%)} & $3.74\pm 0.07$ & $3.74\pm 0.07$ & $3.51 \pm 0.06$ & $3.62 \pm 0.04$ & $4.43 \pm 0.08$ \\
        \multicolumn{1}{l}{Param (M)} & 2.3 & 2.3 & 2.6 & 2.6 & 2.7\\
        \multicolumn{1}{l}{Multiply-Add (M)} & 386 & 386 & 455 & 449 & 444\\
    \bottomrule
    \end{tabular}
\end{table}

\end{document}

%% file: math_commands.tex
\usepackage{amsmath,amsfonts,bm}

\def\eqref#1{equation~\ref{#1}}

\def\1{\bm{1}}

\def\rvu{{\mathbf{i}}}

\def\rvu{{\mathbf{u}}}

\def\rvy{{\mathbf{y}}}
\def\rvz{{\mathbf{z}}}

\def\rmA{{\mathbf{A}}}

\DeclareMathAlphabet{\mathsfit}{\encodingdefault}{\sfdefault}{m}{sl}
\SetMathAlphabet{\mathsfit}{bold}{\encodingdefault}{\sfdefault}{bx}{n}

\newcommand{\R}{\mathbb{R}}

\DeclareMathOperator*{\argmin}{arg\,min}

%% file: latexdefs.tex
\usepackage{amsmath,epsfig}
\usepackage{amssymb}
\usepackage{amsthm}
\usepackage{algorithm}
\usepackage{algpseudocode}
\usepackage{tikz}
\usepackage{tabu}
\usepackage{graphicx}
\usepackage{subfig}
\usepackage{epstopdf}
\usepackage{epsfig}
\usepackage{booktabs}
\usepackage[flushleft]{threeparttable}
\usetikzlibrary{tikzmark,calc}
\usepackage{float}
\usepackage{multirow}
\usepackage{hyperref}
\usepackage{verbatim}
\usepackage{wrapfig}
\usepackage{appendix}
\usepackage{rotating}
\usepackage{mathtools}
\usepackage{bm}
\usepackage{enumitem}
\usepackage{standalone}
\usepackage{pgfplots}
\usepackage{pgfplotstable}
\numberwithin{equation}{section} 
\newtheorem{theorem}{Theorem}[section]                   

\newtheorem{definition}[theorem]{Definition}

\usepackage{microtype}

\def\R{{\mathbb{R}}}

\pgfdeclarelayer{background}
\pgfdeclarelayer{foreground}
\pgfsetlayers{background,main,foreground}

\tikzstyle{sensor}=[draw, fill=blue!20, text width=4em, 
    text centered, minimum height=2.5em]
\tikzstyle{normalCell}=[draw, fill=orange!20, text width=3.2em, 
    text centered, minimum width=2.0em, minimum height=1.4em, 
    rounded corners]
\tikzstyle{conv}=[draw, fill=blue!20, text width=4.2em, 
    text centered, minimum width=2.0em, minimum height=1.4em, rounded corners]
\tikzstyle{operation}=[draw, fill=white!20, text width=3.2em, 
    text centered, minimum width=2.0em, minimum height=1.4em, rounded corners]
\tikzstyle{cell_in}=[draw, fill=green!20, text width=3.0em, 
    text centered, minimum width=2.5em, minimum height=2.3em]
\tikzstyle{cell_out}=[draw, fill=yellow!20, text width=3.0em, 
    text centered, minimum width=2.5em, minimum height=2.3em]
\tikzstyle{square}=[draw, fill=blue!20, minimum size=2em]
\tikzstyle{choice}=[draw, fill=green!20, text width=1.6em, 
    text centered, minimum width=2.0em, minimum height=1.4em, rounded corners]
\tikzstyle{customNode}=[draw, fill=purple!25, text width=3.2em, 
    text centered, minimum width=2.0em, minimum height=1.4em, rounded corners]
\tikzstyle{ann} = [above, text width=5em, text centered]
\tikzstyle{wa} = [sensor, text width=10em, fill=red!20, 
    minimum height=6em, rounded corners, drop shadow]
\tikzstyle{sc} = [sensor, text width=13em, fill=red!20, 
    minimum height=10em, rounded corners, drop shadow]
\tikzstyle{texto} = [above, text width=6em, text centered]
\tikzstyle{oneShotModel}=[draw, fill=white!10, text width=3.0em, 
    text centered, minimum width=2.0em, minimum height=6.0em]
\tikzstyle{sampleNode}=[draw, fill=white!10, text width=5.0em, 
    text centered, minimum width=2.0em, minimum height=1.0em]
\tikzstyle{CSNode}=[draw, fill=white!10, text width=4.0em, 
    text centered, minimum width=2.0em, minimum height=6.0em]
\tikzstyle{restrictionNode}=[draw, fill=white!10, text width=5.0em, 
    text centered, minimum width=2.0em, minimum height=1.0em]

\newcommand{\backgrounda}[6]{%
  \begin{pgfonlayer}{background}
    \path (#1.west |- #2.north)+(-0.2,1.1) node (a1) {};
    \path (#3.east |- #4.south)+(+0.2,-0.25) node (a2) {};
    \path[fill=#5!20,rounded corners, draw=black!50, dashed]
      (a1) rectangle (a2);
    \path (a1.west |- a1.north)+(2.5, -0.8) node (u1)[texto]
      {\large #6};
  \end{pgfonlayer}}
  
\newcommand{\backgroundb}[6]{%
  \begin{pgfonlayer}{background}
    \path (#1.west |- #2.north)+(-0.14,1.157) node (a1) {};
    \path (#3.east |- #4.south)+(+0.5,-0.25) node (a2) {};
    \path[fill=#5!20,rounded corners, draw=black!50, dashed]
      (a1) rectangle (a2);
    \path (a1.west |- a1.north)+(5.2, -0.8) node (u1)[texto]
      {\large #6};
  \end{pgfonlayer}}


%% file: diagram.tex
\resizebox{\linewidth}{!}{
\begin{tikzpicture}
    \fontfamily{cmss}
    \tikzstyle{every node}=[font=\normalsize] 
    \node (output) [operation] {Output};
    \path (output.north)+(0, 0.7) node (soft) [operation] {Softmax};
    \path (soft.north)+(0, 0.7) node (cell3) [normalCell] {Cell3};
    \path (cell3.north)+(0, 0.7) node (cell2) [normalCell] {Cell2};
    \path (cell2.north)+(0, 0.7) node (cell1) [normalCell] {Cell1};
    \path (cell1.north)+(0, 0.7) node (stem2) [operation] {Stem2};
    \path (stem2.north)+(0, 0.7) node (stem1) [operation] {Stem1};
    \path (stem1.north)+(0, 0.7) node (image) [operation] {Image};
    
    \path [draw, ->, line width = 0.4mm] (soft.south) -- node [above] {} (output);
    \path [draw, ->, line width = 0.4mm] (cell3.south) -- node [above] {} (soft);
    \path [draw, ->, line width = 0.4mm] (cell2.south) -- node [above] {} (cell3);
    \path [draw, ->, line width = 0.4mm] (cell1) to [out=340, in=20] (cell3);
    \path [draw, ->, line width = 0.4mm] (cell1.south) -- node [above] {} (cell2);
    \path [draw, ->, line width = 0.4mm] (stem2) to [out=340, in=20] (cell2);
    \path [draw, ->, line width = 0.4mm] (stem2.south) -- node [above] {} (cell1);
    \path [draw, ->, line width = 0.4mm] (stem1) to [out=340, in=20] (cell1);
    \path [draw, ->, line width = 0.4mm] (image) to [out=340, in=20] (stem2);
    \path [draw, ->, line width = 0.4mm] (image.south) -- node [above] {} (stem1);
    
    \draw[dashed] (image.east)+(0.7, 0.3) -- +(0.7, -7.0); 
    
    \path (image.east) + (1.6, 0) node (previousCell1) [operation] {Cell$_{k-2}$};
    \path (image.east) + (5.3, 0) node (previousCell2) [operation] {Cell$_{k-1}$};
    \path (previousCell1.south) + (0, -1.5) node (choice1) [choice] {C1};
    \path (previousCell1.south) + (1.2, -1.5) node (choice2) [choice] {C2};
    \path (choice1.south) + (0.6, -1.0) node (node1) [customNode] {Node1};
    \path (previousCell1.south) + (2.5, -2.5) node (choice3) [choice] {C3};
    \path (previousCell1.south) + (3.7, -2.5) node (choice4) [choice] {C4};
    \path (node1.south) + (0, -0.7) node (choice5) [choice] {C5};
    \path (choice1.south) + (3, -2.5) node (node2) [customNode] {Node2};
    \path (image.east) + (3.4, -6) node (concat) [operation] {Concat.};
    
    \path [draw, ->, line width = 0.4mm] (previousCell1.south) to [out=270, in=90] (choice1);
    \path [draw, ->, line width = 0.4mm] (previousCell1) to [out=320, in=100] (choice3);
    \path [draw, ->, line width = 0.4mm] (previousCell2.south) to [out=270, in=90] (choice4);
    \path [draw, ->, line width = 0.4mm] (previousCell2) to [out=240, in=30] (choice2);
    \path [draw, dashed, ->, line width = 0.4mm] (choice1) to [out=260, in=120] (node1);
    \path [draw, dashed, ->, line width = 0.4mm] (choice2) to [out=280, in=60] (node1);
    \path [draw, dashed, ->, line width = 0.4mm] (choice3) to [out=260, in=110] (node2);
    \path [draw, dashed, ->, line width = 0.4mm] (choice4) to [out=280, in=70] (node2);
    \path [draw, ->, line width = 0.4mm] (node1) to [out=270, in=90] (choice5);
    \path [draw, dashed, ->, line width = 0.4mm] (choice5) to [out=330, in=140] (node2);
    \path [draw, ->, line width = 0.4mm] (node1) to [out=180, in=90] (2.4, 2.7) to [out=270, in=140] (concat.west); 
    \path [draw, ->, line width = 0.4mm] (node2) to [out=230, in=30] (concat.east);
    \draw [->, line width = 0.4mm] (concat.south) to (4.08, -0.2);
    

    \backgrounda{choice1}{choice1}{choice4}{concat}{orange}{\textbf{Cell}}; 
    \draw[dashed] (previousCell2.east)+(0.20, 0.3) -- +(0.20, -7.0);
    
    \path (previousCell2.east) + (5.5, -1.8) node (concat2) [operation] {Input.};
    \path (concat2) + (0, -1.5) node (identity) [conv] {Identity};
    \path (concat2) + (-2, -1.5) node (conv5x5_1) [conv] {5x5};
    \path (concat2) + (-2, -2.5) node (conv5x5_2) [conv] {5x5};
    \path (concat2) + (-4, -1.5) node (conv3x3_1) [conv] {3x3};
    \path (concat2) + (-4, -2.5) node (conv3x3_2) [conv] {3x3};
    \path (concat2) + (2, -1.5) node (maxPool) [conv] {Max Pool};
    \path (concat2) + (4, -1.5) node (avgPool) [conv] {Avg Pool};
    \path (concat2.south) + (0, -4) node (sum) [operation] {Sum};
    
    
    \draw [dashed, ->, line width = 0.4mm] (concat2) to [out=200, in=40] node [above] {\normalsize $\alpha_{i}$} (conv3x3_1);
    \path [draw, ->, line width = 0.4mm] (conv3x3_1) to [out=270, in=90] (conv3x3_2);
    \path [draw, dashed, ->, line width = 0.4mm] (concat2) to [out=220, in=60] node [below, pos=0.4] {\normalsize $\alpha_{i+1}$} (conv5x5_1);
    \path [draw, ->, line width = 0.4mm] (conv5x5_1) to [out=270, in=90] (conv5x5_2);
    \draw [dashed, ->, line width = 0.4mm] (concat2) to [out=270, in=90] node [] {\normalsize $\alpha_{i+2}$} (identity);
    \path [draw, dashed, ->, line width = 0.4mm] (concat2) to [out=320, in=130] node [above, pos=0.8] {\normalsize $\alpha_{i+3}$} (maxPool);
    \path [draw, dashed, ->, line width = 0.4mm] (concat2) to [out=340, in=140] node [above] {\normalsize $\alpha_{i+4}$} (avgPool);
    \path [draw, dashed, ->, line width = 0.4mm] (conv3x3_2) to [out=270, in=180] (sum);
    \path [draw, dashed, ->, line width = 0.4mm] (conv5x5_2) to [out=270, in=170] (sum);
    \path [draw, dashed, ->, line width = 0.4mm] (identity) to [out=270, in=90] (sum);
    \path [draw, dashed, ->, line width = 0.4mm] (maxPool) to [out=270, in=10] (sum);
    \path [draw, dashed, ->, line width = 0.4mm] (avgPool) to [out=270, in=0] (sum);
    \draw [->, line width = 0.4mm] (sum.south) to (12.175, -0.2);
    
    \backgroundb{conv3x3_1}{concat2}{avgPool}{sum}{green}{\textbf{Choice $i$}};
\end{tikzpicture}}